\pdfoutput=1

\documentclass[11pt]{article}
\usepackage[preprint]{acl}
\usepackage{times}
\usepackage{float}
\usepackage{latexsym}
\usepackage{comment}
\usepackage{booktabs}
\usepackage[T1]{fontenc}
\usepackage[utf8]{inputenc}
\usepackage{microtype}
\usepackage{inconsolata}
\usepackage{graphicx}

\newlength{\myMheight}
\newcommand{\hf}{\includegraphics[height=\myMheight]{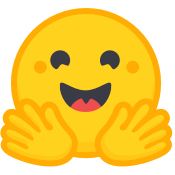}}

\title{Recipient Profiling:\\ Predicting Characteristics from Messages}

\author{
    \textbf{Martín Bórquez\textsuperscript{1}}, 
    \textbf{Mikaela Keller\textsuperscript{1}},
     \textbf{Michael Perrot\textsuperscript{1}}, 
     \textbf{Damien Sileo\textsuperscript{1}}, 
    \\
     \textsuperscript{1} Univ. Lille, Inria, CNRS, Centrale Lille, UMR 9189 - CRIStAL, F-59000 Lille, France
     \\
     \small{
        \texttt{\href{mailto:borquezmartin@uc.cl}{borquezmartin@uc.cl},
        \href{mailto:mikaela.keller@inria.fr}{mikaela.keller@inria.fr},
        \href{mailto:michael.perrot@inria.fr}{michael.perrot@inria.fr},
        \href{mailto:damien.sileo@inria.fr}{damien.sileo@inria.fr}
         }
    }
    }

\begin{document}
\maketitle

\begin{abstract}
It has been shown in the field of Author Profiling that texts may inadvertently reveal sensitive information about their authors, such as gender or age. This raises important privacy concerns that have been extensively addressed in the literature, in particular with the development of methods to hide such information. We argue that, when these texts are in fact messages exchanged between individuals, this is not the end of the story. Indeed, in this case, a second party, the intended recipient, is also involved and should be considered. In this work, we investigate the potential privacy leaks affecting them, that is we propose and address the problem of Recipient Profiling. We provide empirical evidence that such a task is feasible on several publicly
accessible datasets\footnote{\href{https://huggingface.co/datasets/sileod/recipient_profiling}{[data:HF-datasets \hf]}}. Furthermore, we show that the learned models can be transferred to other datasets, albeit with a loss in accuracy.
\end{abstract}

\section{Introduction}

Written and spoken exchanges are essential means to share information between individuals \cite{CornbleetCarter2002, Fer2018}. They usually involve an author that communicates with some intended recipients such as an audience or a single individual. For example, journalists write articles for their readers while friends exchange messages in private conversations. It is noteworthy that, beyond their initial and intended content, these exchanges might also unintentionally and inadvertently reveal some features of the involved parties such as their genders, ages, or personality traits \citep{Rangel2013,El2014}. This can be due to the presence of direct identifiers, such as pronoun or titles of address like ``sir'' or ``madam'', but also to more subtle and indirect identifiers, such as stylistic adaptations made by authors depending on their audience \citep{Giles2007,giles2023}.

Identifying these features can be of significant interest as they provide clues to understand how individuals perceive both themselves and others, and how these perceptions might be shared inadvertently. However, this also raises questions regarding the privacy of the parties involved in the exchanges since these features might be of a potentially sensitive nature. It is thus natural to wonder to which extent it is possible to automatically infer such information from texts alone and whether it is possible to hide it. This led to the development of tasks in the natural language processing and privacy preserving machine learning communities. In particular, in author profiling the goal is to infer sensitive features, such as gender, age, or personality traits, from the author of a text \citep{Rangel2013,rangel2013overview,El2014,koolen2017,onikoyi2023,Ouni2023}. Similarly, several approaches have been proposed to hide such sensitive information \citep{mamede2016automated,lison2021anonymisation,loiseau2024tarot,fisher-etal-2024-jamdec}. However, to the best of our knowledge, the development of these new tasks has been solely focused on authors while recipients have been ignored. We believe that this is an important oversight and, in this work, we address it by proposing a new task that we call Recipient Profiling.

Our contributions are twofold. First, we formalize the problem of Recipient Profiling by taking inspiration from Author Profiling. Roughly speaking, we define it as the task of predicting one or several sensitive attributes of a recipient from one or several pieces of text they received. Second, we provide proof-of-concept experiments that show the feasibility of this new task on several datasets. More precisely, we empirically show that three recent language models obtain a higher-than-chance probability of correctly predicting the gender of recipients on three different datasets. Furthermore, we show some amount of transferability as the models trained on any given dataset perform better than chance on the others.

\section{Related work}

To the best of our knowledge, the task of Recipient Profiling was not formalized before in the literature. Nevertheless, we identify several works that seem particularly relevant.

\paragraph{Author Profiling.} Author Profiling has received some attention in the literature and we can distinguish two main families of approaches. On the one hand, the early works relied on linguistic patterns and hand-crafted features to determine attributes such as age, gender, and personality traits \citep{Rangel2013,El2014,koolen2017}. On the other hand, recent advancements in language models provided new opportunities as they enabled analyses across various domains at a larger scale. They expanded the possibilities for more comprehensive studies thanks to their ability to uncover subtle and complex relationships between the text and the characteristics of the authors. For example, \citet{koolen2017} present early work on author gender identification, emphasizing the impact of the target audience on the writing style of authors. Similarly, \citet{onikoyi2023} present a gender identification method for Twitter users using NLP and machine learning techniques. As a last example, \citet{Ouni2023} examine current author profiling methods for identifying demographic and psychological traits from social network texts, evaluating their effectiveness. In this paper, we also rely on the capabilities of recent text encoders in our proof-of-concept experiments as we show that a simple fine-tuning step is sufficient to predict the gender of recipients.

\paragraph{Language Accommodation.} Authors are known to adapt their writing style to suit their audience, a phenomenon termed Language Accommodation.
This communication theory has been explored across various cultural contexts, languages, social groups, practical applications, and disciplines \cite{giles2023}. Additionally, it has shaped the creation of targeted interventions and training in sectors such as healthcare, education, finance, and business \cite{watson2020, frey2021, brau2022, ayeni2021}. An extreme case of such accommodation has been investigated by \citet{tic} who explore a language-modeling based approach to reveal gender bias in tennis post-match interviews. They showed that journalists ask male athletes more game-focused questions compared to female athletes. In this paper, we address a different problem, namely Recipient Profiling. However, the two questions are connected and could enrich one another. For example, known accommodation patterns, that is text adaptions identified in the literature as due to the recipient characteristics, could serve as a basis to design better models for Recipient Profiling. Likewise, the models learned to perform recipient profiling could help practitioners uncover new and subtle accommodation patterns.

\section{Recipient Profiling \label{sec:recpro}}
In this section, we propose a formalization of the task of Recipient Profiling.

Let $a \in \mathcal{A}$ denote the author of a text message and $r \in \mathcal{R}$ its recipient in any written or spoken exchange. We denote by $\mathcal{U}_{\mathcal{A} \rightarrow \mathcal{R}}$ the set of all messages that could be exchanged between the authors in $\mathcal{A}$ and the recipients in $\mathcal{R}$. We further denote by $\mathcal{U}_{a \rightarrow r}$ the set of all utterances that author $a$ could potentially address to recipient $r$ and by $U_{a\rightarrow r} = \left\{u_{a\rightarrow r}^i\right\}_{i=1}^n \subseteq \mathcal{U}_{a \rightarrow r}$ a subset of $n$ utterances that author $a$ sent to recipient $r$. 
We assume that each recipient is associated with a sensitive feature $s_r$ which can take values in a set $\mathcal{S}$. This feature could, for example, be the gender or the age of the recipient. The main goal in Recipient Profiling is to predict the value of the sensitive attribute associated with a recipient from the received message. More formally, the task is to learn a classifier $f : \mathcal{U}_{\mathcal{A} \rightarrow \mathcal{R}} \rightarrow \mathcal{S}$, that is a classifier such that $f(u_{a \rightarrow r})$ is equal to $s_r$.

If the texts exchanged between authors and recipients come from conversations, the utterances might be too short to infer anything, for example ``Hi'', or ``How are you?''. In this case, one can assume that each $u_{a \rightarrow r}$ is in fact a concatenation of several messages to reach a minimal base length.

\section{Experimental setup}
In this section, we present the datasets, models, and evaluation metrics that we use to demonstrate the feasibility of recipient profiling.
We selected three conversation datasets that include the gender of the recipients.

\paragraph{Switchboard Dialog Act Corpus (SWDA)} \citep{swda}. It contains transcripts of telephone conversations about different topics such as childcare, recycling, and news media. Some utterances in this dataset were very short. Hence, as a pre-processing step, we concatenated  segments from the same conversations until we reached a limit of 1,000 characters (as described in the previous section). This decreased the number of available utterances from 122,646 to 9,030. We also filtered out the meta-data tags. 

\paragraph{Movie Dialog Corpus (MDC)} \citep{mdc}. It contains dialogues from a large variety of movie scripts. For this dataset, the only pre-processing involved removing utterances associated with characters whose gender was unknown.

\paragraph{Tennis Interviews Corpus (TIC)} \citep{tic}. It is a collection of transcripts of interviews with professional tennis players after their singles matches. In this dataset we only have access to the gender of the interviewed tennis player and not the journalist. Hence, we only use the questions of the journalists and discard the answers. We did not perform any other specific pre-processing step on the texts.

The datasets we selected presented some degree of imbalance between the two classes. 
To achieve balance, we randomly subsampled the over-represented class, reducing its size to match that of the under-represented class \citep{Kubat1997}. 
We split the datasets into train, validation, and test sets with a recipient-based group split, so that no recipient appears in multiple splits, to prevent overfitting.
We construct random splits using $80\%$ of recipients in the train set and $20\%$ distributed in the test and validation sets.
We provide detailed statistics on our datasets before and after the pre-processing steps in Table~\ref{tab:dataset-description} in the Appendix.

To capture the linguistic patterns relevant for recipients gender prediction, we selected three pre-trained text classification models that are well-known for their effectiveness in author representation \citep{onikoyi2023}. The first one is \textbf{BERT} \citep{bert}. It is often employed for text classification due to its encoder architecture. The second model we selected is \textbf{MPNet} \citep{mpnet} that is known for its performance in sentence embedding. Lastly, we settled for a third model named \textbf{DeBERTa} \citep{deberta} specifically trained for Natural Language Inference \citep{sileo2023deberta}. We added to all the models a classification head consisting in a single linear layer predicting the gender of the recipient. We fine-tune all the weights in all our experiments, using Adam \cite{diederik2014adam}
 with a learning rate of 2e-5 over 3 epochs. 

 We focus our evaluation on \textbf{balanced accuracy}, which is the average accuracy across the classes \citep{Brodersen2010}. In the appendix, we also report results in terms of precision, recall, F1-score, and accuracy.

\begin{figure}[]
    \centering
    \includegraphics[scale=0.50]{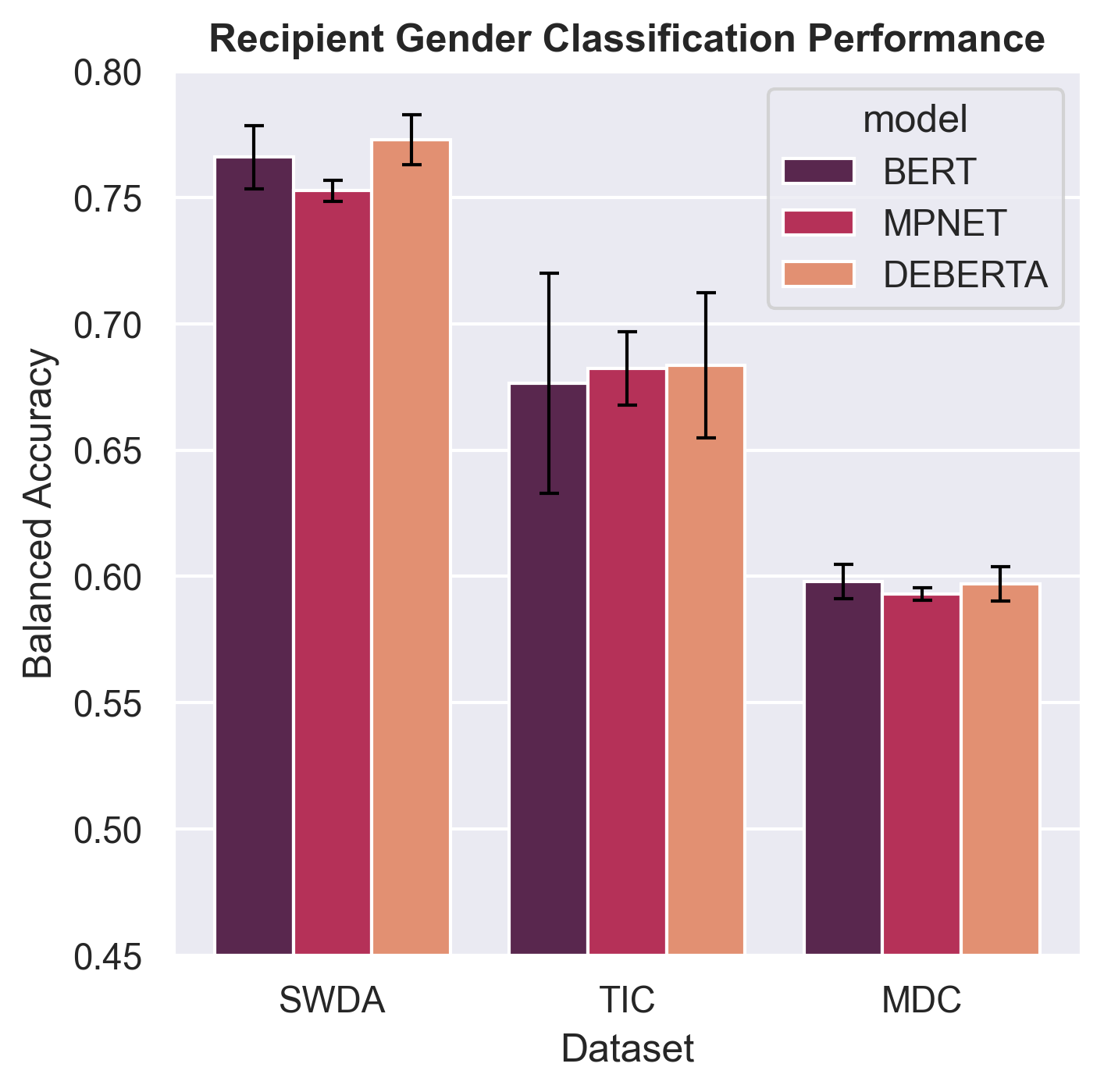}
    \caption{\textbf{Performance of fine-tuned models for recipient gender classification} in terms of balanced accuracy. The barplot shows the performance of each model when trained and tested within the same domain. The error bars represent the standard deviation of the measurements, calculated after running each model with three different seeds.}
    \label{fig:rp-performance}
\end{figure}

\section{Results and Discussion}

In this section, we present our main results that correspond to balanced accuracies averaged over three different runs and the corresponding standard deviations. We start by presenting results on the vanilla recipient profiling task. Second, we present a transfer experiments between datasets. Third, we analyze the difference in performance between genders. Finally, we provide a short analysis showing that the models potentially capture relatively different patterns. 

\subsection{Performance of Fine-Tuned Models}

Figure \ref{fig:rp-performance} illustrates the performance of the fine-tuned models on our different datasets. The main take-away is that we achieve better-than-chance accuracy on all the problems we considered, showing that it is indeed possible to predict the gender of a recipient from a message alone. Note that accuracy, recall, and F1-scores all exhibit similar trends as balanced accuracy as can be seen in Table \ref{tab:recipient-performance-detail} in the appendix.

The findings demonstrate that significant results can be achieved in predicting recipient gender from text messages. All fine-tuned models exceeded baseline performance in this task. Pre-trained text encoders, widely used in other contexts, also performed well, indicating that recipient profiling using various machine learning techniques warrants further exploration.

\subsection{Cross-Dataset Model Transferability}
We assessed how well our models generalize across different domains, understanding that variations may stem from domain-specific vocabulary or stylistic nuances. We evaluated each fine-tuned model on different datasets to observe whether they learned transferable patterns. Most models achieved classification results better than chance, that is balanced accuracies in the interval $[0.51, 0.58]$, even in varied contexts. This could suggest that our models capture patterns that remain valid from one dataset to the other.The full results can be found in Figure \ref{fig:recipient_transfer} in the Appendix \ref{sec:results_appendix}.

\subsection{Accuracy Difference by Gender}
We evaluate the fine-tuned models accuracy by recipient gender. Using three different seeds, we found that, on average, the models predicted female recipients more accurately than male ones by 6.85\% (details in Figure \ref{fig:gender-diff} in Appendix \ref{sec:results_appendix}). The difference varied significantly by dataset and model. Whether this difference is due to a larger amount of direct identifiers for females compared to males or to other factors, for example indirect identifiers, is still an open question. We think that this is an interesting avenue for future work.

\subsection{Models Agreement Analysis}
To assess whether models capture consistent or varied patterns across datasets, we evaluated the agreement in predictions among different models. We compared correct prediction agreement, while accounting for random agreements, for each fine-tuned model within the same context (further details in Figure \ref{fig:kappa-agreement} in the Appendix \ref{sec:models-agreement}
). The analysis aligned with prior performance observations, with lower balanced accuracy models showing reduced agreement compared to higher accuracy models. Although models with improved accuracy exhibited greater agreement, it remained weak (kappa coefficient between0.46 and 0.61 \cite{McHugh2012}. This indicates that while some patterns are interpreted in a similar way, others are not. These insights suggest that combining different datasets could be worth exploring in further studies to improve model performance in recipient profiling.

\section{Conclusion}
In this work, we formalized a new natural language processing task called recipient profiling and showed its feasibility using three different text encoders (BERT, MPNet and DeBERTa) and three datasets (SWDA, MDC, TIC).

We believe that this opens up different research avenues. First, it would be interesting to apply explainability techniques to identify the nature of the patterns leading to correct predictions. These could then be analyzed under the lens of known linguistic phenomenon, for example to unveil the presence of direct and indirect identifiers that could be linked to accommodation. Second, the feasibility of recipient profiling constitutes a new privacy risk in textual data. It means that new mitigation techniques will need to be developed in the future, for example approaches to write texts that are neutral with respect to their recipient. Third, it would be interesting to go beyond text and explore other data modalities where different kind of clues could be helpful to profile recipients. For example, we could consider speech data and take into account the prosody of the author.

\section{Limitations}

There is two main aspects to our work, the formalization of Recipient Profiling and some experiments to show the feasibility of the task. We identify some limitations in both parts. On the one hand, our experiments are preliminary and only serve as a proof-of-concept. Considering wider empirical investigations, with other models beyond off-the-shelf text encoders and other datasets, could give a better understanding of the problem we are unveiling in this paper. Furthermore, our analysis of the results remains quite superficial. Indeed, we did not investigate the impact of utterance length on the performance of our models or whether their predictions were due to direct or indirect identifiers. Similarly, we did not measure the impact of stylometric features that are known to play a role in author profiling \citep{fabien-etal-2020-bertaa}.

On the other hand, we only consider a basic Recipient Profiling formalism in Section \ref{sec:recpro} while other interesting settings could be relevant.

\paragraph{Multiple sensitive features.} In some cases, each recipient will possess multiple sensitive features such as gender, age, and ethnicity. In this case, we can either use intersectionality where $\mathcal{S}$ becomes the set of all possible combinations of sensitive features or assume that there exists sensitive sets $\mathcal{S}_1, \ldots, \mathcal{S}_K$.

\paragraph{Fuzzy sensitive features.} Another setting of interest is when the recipients do not exactly belong to one clear-cut group identified by the sensitive feature but, instead, there is some permeability between the groups that the feature of interest defines and recipients belong to them to various degrees. This is for example the case when we consider personality traits. In this case, instead of predicting a single attribute, the learned classifier may output a probability distribution over all the possible output, that is $f : \mathcal{U}_{\mathcal{A} \rightarrow \mathcal{R}} \rightarrow \mathcal{P}(\mathcal{S})$ where $\mathcal{P}(\mathcal{S})$ represents the set of possible probability distributions over $\mathcal{S}$.

\paragraph{Taking authors into account.} In our basic formalization of Recipient Profiling, we completely ignore the author characteristics. However, it would be interesting to jointly consider Author and Recipient Profiling as these two tasks could complement one another. This could be particularly relevant in cases where the same author and recipient pair continuously exchange messages.

\section{Ethical Considerations}

The possibility of Recipient Profiling raises significant ethical concerns, particularly regarding privacy and potential discriminatory implications. Our research primarily focused on the privacy dimensions, recognizing the complex challenges in definitively separating predictive signals from underlying societal biases.
Privacy emerges as a critical concern. Our experiments demonstrate that message contents can inadvertently reveal sensitive characteristics about recipients, potentially without their consent or awareness. This capability could be misused for unwarranted personal information extraction, raising substantial privacy risks in digital communication contexts.
The potential for bias analysis is equally important. While our current work does not comprehensively disentangle whether predictions stem from genuine linguistic markers or stereotypical patterns, we acknowledge this as a crucial avenue for future research. Predicting recipient characteristics could potentially help identifying unintended stereotypes or patronizing language \citep{perez-almendros-etal-2022-semeval} while writing text for a specific audience.

\bibliography{custom}

\begin{thebibliography}{31}
\providecommand{\natexlab}[1]{#1}

\bibitem[{Ayeni(2021)}]{ayeni2021}
Bartholomew Ayeni. 2021.
\newblock \href {https://doi.org/10.7575/aiac.ijalel.v.10n.1p.80} {Language choices and its effect in a culturally diversified nigeria business places: Adopting giles’ communication accommodation theory}.
\newblock \emph{International Journal of Applied Linguistics and English Literature}.

\bibitem[{Brau et~al.(2022)Brau, Cicon, and Owen}]{brau2022}
James~C. Brau, James Cicon, and Stephen~R. Owen. 2022.
\newblock \href {https://doi.org/10.3390/ijfs10020025} {A textual analysis of logograms in chinese ipo roadshows: How agreement between investors and management relates to pricing and performance}.
\newblock \emph{International Journal of Financial Studies}, 10(2):25.

\bibitem[{Brodersen et~al.(2010)Brodersen, Ong, Stephan, and Buhmann}]{Brodersen2010}
K.H. Brodersen, C.S. Ong, K.E. Stephan, and J.M. Buhmann. 2010.
\newblock The balanced accuracy and its posterior distribution.
\newblock In \emph{Proceedings of the 20th International Conference on Pattern Recognition}, pages 3121--3124.

\bibitem[{Cornbleet and Carter(2002)}]{CornbleetCarter2002}
Sandra Cornbleet and Ronald Carter. 2002.
\newblock \emph{The Language of Speech and Writing}.
\newblock Routledge.
\newblock Creative Commons license (2001). Available at \url{http://ndl.ethernet.edu.et/bitstream/123456789/8674/1/46.pdf.pdf}.

\bibitem[{Danescu-Niculescu-Mizil and Lee(2011)}]{mdc}
Cristian Danescu-Niculescu-Mizil and Lillian Lee. 2011.
\newblock Chameleons in imagined conversations: A new approach to understanding coordination of linguistic style in dialogs.
\newblock In \emph{Proceedings of the Workshop on Cognitive Modeling and Computational Linguistics, ACL 2011}.

\bibitem[{Devlin et~al.(2019)Devlin, Chang, Lee, and Toutanova}]{bert}
Jacob Devlin, Ming-Wei Chang, Kenton Lee, and Kristina Toutanova. 2019.
\newblock \href {https://doi.org/10.18653/v1/N19-1423} {Bert: Pre-training of deep bidirectional transformers for language understanding}.
\newblock In \emph{Proceedings of the 2019 Conference of the North American Chapter of the Association for Computational Linguistics: Human Language Technologies, Volume 1 (Long and Short Papers)}, pages 4171--4186, Minneapolis, Minnesota. Association for Computational Linguistics.

\bibitem[{Diederik(2014)}]{diederik2014adam}
P~Kingma Diederik. 2014.
\newblock Adam: A method for stochastic optimization.
\newblock \emph{(No Title)}.

\bibitem[{El~Manar and Kassou(2014)}]{El2014}
Sara El~Manar and Ismail Kassou. 2014.
\newblock Authorship analysis studies: A survey.
\newblock \emph{International Journal of Computer Applications}, 86(12):22--29.

\bibitem[{Fabien et~al.(2020)Fabien, Villatoro-Tello, Motlicek, and Parida}]{fabien-etal-2020-bertaa}
Ma{\"e}l Fabien, Esau Villatoro-Tello, Petr Motlicek, and Shantipriya Parida. 2020.
\newblock \href {https://aclanthology.org/2020.icon-main.16} {{B}ert{AA} : {BERT} fine-tuning for authorship attribution}.
\newblock In \emph{Proceedings of the 17th International Conference on Natural Language Processing (ICON)}, pages 127--137, Indian Institute of Technology Patna, Patna, India. NLP Association of India (NLPAI).

\bibitem[{Fer(2018)}]{Fer2018}
Simona Fer. 2018.
\newblock \href {https://doi.org/10.2139/ssrn.3128115} {Verbal communication as a two-way process in connecting people}.
\newblock SSRN Electronic Journal.
\newblock Available at SSRN.

\bibitem[{Fisher et~al.(2024)Fisher, Lu, Jung, Jiang, Harchaoui, and Choi}]{fisher-etal-2024-jamdec}
Jillian Fisher, Ximing Lu, Jaehun Jung, Liwei Jiang, Zaid Harchaoui, and Yejin Choi. 2024.
\newblock \href {https://doi.org/10.18653/v1/2024.naacl-long.87} {{JAMDEC}: Unsupervised authorship obfuscation using constrained decoding over small language models}.
\newblock In \emph{Proceedings of the 2024 Conference of the North American Chapter of the Association for Computational Linguistics: Human Language Technologies (Volume 1: Long Papers)}, pages 1552--1581, Mexico City, Mexico. Association for Computational Linguistics.

\bibitem[{Frey and Lane(2021)}]{frey2021}
Timothy~K. Frey and Derek~R. Lane. 2021.
\newblock \href {https://doi.org/10.1080/03634523.2021.1903521} {Cat in the classroom: A multilevel analysis of students’ experiences with instructor nonaccommodation}.
\newblock \emph{Communication Education}, 70(3):223--246.

\bibitem[{Fu et~al.(2016)Fu, Danescu-Niculescu-Mizil, and Lee}]{tic}
Liye Fu, Cristian Danescu-Niculescu-Mizil, and Lillian Lee. 2016.
\newblock Tie-breaker: Using language models to quantify gender bias in sports journalism.
\newblock In \emph{Proceedings of the IJCAI workshop on NLP meets Journalism}.

\bibitem[{Giles et~al.(2023)Giles, Edwards, and Walther}]{giles2023}
Howard Giles, America~L. Edwards, and Joseph~B. Walther. 2023.
\newblock \href {https://doi.org/10.1016/j.langsci.2023.101571} {Communication accommodation theory: Past accomplishments, current trends, and future prospects}.
\newblock \emph{Language Sciences}, 99:101571.
\newblock Accessed: 2024-08-14.

\bibitem[{Giles and Ogay(2007)}]{Giles2007}
Howard Giles and Tania Ogay. 2007.
\newblock Communication accommodation theory.
\newblock In B.~B. Whaley and W.~Samter, editors, \emph{Explaining Communication: Contemporary Theories and Exemplars}, pages 293--310. Lawrence Erlbaum Associates Publishers.

\bibitem[{He et~al.(2021)He, Liu, Gao, and Chen}]{deberta}
Pengcheng He, Xiaodong Liu, Jianfeng Gao, and Weizhu Chen. 2021.
\newblock \href {https://arxiv.org/abs/2006.03654} {Deberta: Decoding-enhanced bert with disentangled attention}.
\newblock \emph{Preprint}, arXiv:2006.03654.

\bibitem[{Koolen and van Cranenburgh(2017)}]{koolen2017}
Corina Koolen and Andreas van Cranenburgh. 2017.
\newblock \href {https://doi.org/10.18653/v1/W17-1602} {These are not the stereotypes you are looking for: Bias and fairness in authorial gender attribution}.
\newblock In \emph{Proceedings of the First {ACL} Workshop on Ethics in Natural Language Processing}, pages 12--22, Valencia, Spain. Association for Computational Linguistics.

\bibitem[{Kubat and Matwin(1997)}]{Kubat1997}
Miroslav Kubat and Stan Matwin. 1997.
\newblock Addressing the curse of imbalanced training sets: One-sided selection.
\newblock In \emph{Proceedings of the Fourteenth International Conference on Machine Learning (ICML '97)}, pages 179--186, San Francisco, CA, USA. Morgan Kaufmann.

\bibitem[{Lison et~al.(2021)Lison, Pil{\'a}n, S{\'a}nchez, Batet, and {\O}vrelid}]{lison2021anonymisation}
Pierre Lison, Ildik{\'o} Pil{\'a}n, David S{\'a}nchez, Montserrat Batet, and Lilja {\O}vrelid. 2021.
\newblock Anonymisation models for text data: State of the art, challenges and future directions.
\newblock In \emph{Proceedings of the 59th Annual Meeting of the Association for Computational Linguistics and the 11th International Joint Conference on Natural Language Processing (Volume 1: Long Papers)}, pages 4188--4203.

\bibitem[{Loiseau et~al.(2024)Loiseau, Sileo, Riquet, Meyer, and Tommasi}]{loiseau2024tarot}
Gabriel Loiseau, Damien Sileo, Damien Riquet, Maxime Meyer, and Marc Tommasi. 2024.
\newblock Tarot: Task-oriented authorship obfuscation using policy optimization methods.
\newblock \emph{arXiv preprint arXiv:2407.21630}.

\bibitem[{Mamede et~al.(2016)Mamede, Baptista, and Dias}]{mamede2016automated}
Nuno Mamede, Jorge Baptista, and Francisco Dias. 2016.
\newblock Automated anonymization of text documents.
\newblock In \emph{2016 IEEE congress on evolutionary computation (CEC)}, pages 1287--1294. IEEE.

\bibitem[{McHugh(2012)}]{McHugh2012}
Mary~L. McHugh. 2012.
\newblock Interrater reliability: The kappa statistic.
\newblock \emph{Biochemia Medica (Zagreb)}, 22(3):276--282.

\bibitem[{Onikoyi et~al.(2023)Onikoyi, Nnamoko, and Korkontzelos}]{onikoyi2023}
Babatunde Onikoyi, Nonso Nnamoko, and Ioannis Korkontzelos. 2023.
\newblock \href {https://doi.org/10.1016/j.nlp.2023.100018} {Gender prediction with descriptive textual data using a machine learning approach}.
\newblock \emph{Natural Language Processing Journal}, 4:100018.
\newblock Accessed: 2024-08-12.

\bibitem[{Ouni et~al.(2023)Ouni, Fkih, and Omri}]{Ouni2023}
Sarra Ouni, Faten Fkih, and Mohamed~Nazih Omri. 2023.
\newblock \href {https://doi.org/10.1007/s11042-023-14711-8} {A survey of machine learning-based author profiling from texts analysis in social networks}.
\newblock \emph{Multimedia Tools and Applications}, 82:36653--36686.

\bibitem[{Perez-Almendros et~al.(2022)Perez-Almendros, Espinosa-Anke, and Schockaert}]{perez-almendros-etal-2022-semeval}
Carla Perez-Almendros, Luis Espinosa-Anke, and Steven Schockaert. 2022.
\newblock \href {https://doi.org/10.18653/v1/2022.semeval-1.38} {{S}em{E}val-2022 task 4: Patronizing and condescending language detection}.
\newblock In \emph{Proceedings of the 16th International Workshop on Semantic Evaluation (SemEval-2022)}, pages 298--307, Seattle, United States. Association for Computational Linguistics.

\bibitem[{Rangel and Rosso(2013)}]{Rangel2013}
Francisco Rangel and Paolo Rosso. 2013.
\newblock \href {https://www.semanticscholar.org/paper/Use-of-Language-and-Author-Profiling-%3A-of-Gender-Rangel-Rosso/a4f1b66c7bdaa551c4931fcb9c482f8d5f788db2} {Use of language and author profiling: Identification of gender and age}.
\newblock In \emph{7th International Conference on Corpus Linguistics}.
\newblock Accessed: 2024-07-09.

\bibitem[{Rangel et~al.(2013)Rangel, Rosso, Koppel, Stamatatos, and Inches}]{rangel2013overview}
Francisco Rangel, Paolo Rosso, Moshe Koppel, Efstathios Stamatatos, and Giacomo Inches. 2013.
\newblock Overview of the author profiling task at pan 2013.
\newblock In \emph{CLEF conference on multilingual and multimodal information access evaluation}, pages 352--365. CELCT.

\bibitem[{Sileo(2023)}]{sileo2023deberta}
Damien Sileo. 2023.
\newblock \href {https://arxiv.org/abs/2301.05948} {tasksource: Structured dataset preprocessing annotations for frictionless extreme multi-task learning and evaluation}.
\newblock \emph{arXiv preprint arXiv:2301.05948}.

\bibitem[{Song et~al.(2020)Song, Tan, Qin, Lu, and Liu}]{mpnet}
Kaitao Song, Xu~Tan, Tao Qin, Jianfeng Lu, and Tie-Yan Liu. 2020.
\newblock \href {https://arxiv.org/abs/2004.09297} {Mpnet: Masked and permuted pre-training for language understanding}.
\newblock \emph{Preprint}, arXiv:2004.09297.

\bibitem[{Stolcke et~al.(2000)Stolcke, Coccaro, Bates, Taylor, Van Ess-Dykema, Ries, Shriberg, Jurafsky, Martin, and Meteer}]{swda}
A.~Stolcke, N.~Coccaro, R.~Bates, P.~Taylor, C.~Van Ess-Dykema, K.~Ries, E.~Shriberg, D.~Jurafsky, R.~W. Martin, and M.. Meteer. 2000.
\newblock \href {https://doi.org/10.1162/089120100561737} {Dialogue act modeling for automatic tagging and recognition of conversational speech}.
\newblock \emph{Computational Linguistics - Association for Computational Linguistics}, 26(3):339–373.

\bibitem[{Watson(2020)}]{watson2020}
Brian~M. Watson. 2020.
\newblock \href {https://doi.org/10.1007/978-3-030-41668-3_9} {Communication accommodation theory as an intervention tool to improve interprofessional practice in healthcare}.
\newblock In Louise Mullany, editor, \emph{Professional Communication. Communicating in Professions and Organizations}. Palgrave Macmillan, Cham.

\end{thebibliography}

\onecolumn \appendix

\section{Datasets Details}
\label{sec:dataset-appendix}

Table \ref{tab:dataset-description} provides a detailed description of the datasets used in this paper. It includes information on the number of utterances, authors, recipients, the gender distribution, and the specific details of the experimental splits.

\begin{table*}[h]
  \centering
  \small
  \begin{tabular}{lccc}
    \toprule
    \textbf{Dataset Code}  & \textbf{SWDA} & \textbf{MDC} & \textbf{TIC} \vspace{1mm}\\ 
    \hline
    \textbf{Gender annotation}  & Author and Recipient & Author and Recipient & Only recipient \\
    \textbf{Utterances}  & {\color[HTML]{000000} 122,646} & 288,341 & 81,974 \\
    \textbf{Number of Recipients} & 440 & 8,749 & 358 \\
    \textbf{Female/Male Recipients}  & 220 / 220 & 948 / 2003 & 167 / 191 \\
    \textbf{Balanced Utterances*}  & 9,030 & 46,088 & 75,994 \\
    \textbf{Mean Characters per Balanced Utterances*}  & 926.96 & 53.37 & 103.07 \\
    \textbf{Recipient train/val/test} & 352 / 18 / 70 & 2361 / 118/ 472 & 250 / 33 / 75      \\
    \textbf{Utterances train/val/test} & 5,466 / 516 / 1,578   & 31,924 / 4,233 / 9,931  & 48,920 / 13,698 / 13,376     \\
    \bottomrule
  \end{tabular}
  \caption{\label{tab:dataset-description} Datasets description. Balanced utterances refers to the number of utterances that were used in the experiments after pre-processing to balance the number of male and female author and recipients. In the MDC dataset we had to drop some utterances as the gender information was only available for $2951$ recipients out of $8749$.}
\end{table*}

\section{Experiment Results Details \label{sec:results_appendix}}
\begin{figure}[H]
    \centering
    \includegraphics[scale=0.45]{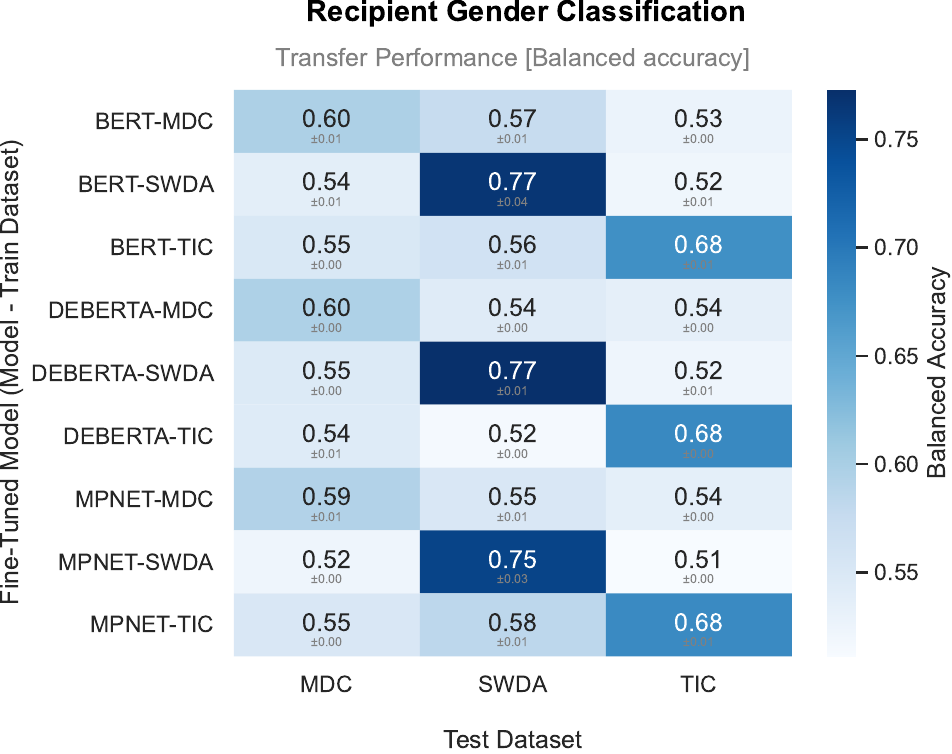}
    \caption{Balanced accuracy transfer performance of fine-tuned models for recipient gender classification. The values represent the mean balanced accuracy, over three seeds for each model, with their respective standard deviations. }
    \label{fig:recipient_transfer}
\end{figure}

\begin{table}[H]
  \centering
  \small
  \begin{tabular}{c l l c c c c c c c}
    \toprule
    \textbf{ID} & \textbf{Model} & \textbf{Dataset} & \textbf{Balanced Accuracy} & \textbf{Accuracy} & \textbf{F1-score} & \textbf{Precision} & \textbf{Recall} & \textbf{Seeds} \\
    \midrule
    1 & BERT & SWDA & 0.7659 & 0.7662 & 0.7633 & 0.7625 & 0.7659 & 3 \\
       &      &      & (0.0435) & (0.0440) & (0.0459) & (0.0461) & (0.0435) &   \\
    2 & MPNET & SWDA & 0.7527 & 0.7612 & 0.7540 & 0.7584 & 0.7527 & 3 \\
       &       &      & (0.0287) & (0.0264) & (0.0292) & (0.0324) & (0.0287) &   \\
    {3} & {DEBERTA} & {SWDA} & {0.7729} & {0.7739} & {0.7707} &{ 0.7704} & {0.7729} & 3 \\
       &         &      & {(0.0145)} & {(0.0168)} & { (0.0167)} & {(0.0175)} & {(0.0145)} &   \\
    4 & BERT & MDC & 0.5980 & 0.5943 & 0.5939 & 0.5991 & 0.5980 & 3 \\
       &      &     & (0.0126) & (0.0135) & (0.0137) & (0.0129) & (0.0126) &   \\
    5 & MPNET & MDC & 0.5930 & 0.5916 & 0.5901 & 0.5942 & 0.5930 & 3 \\
       &       &     & (0.0098) & (0.0107) & (0.0114) & (0.0102) & (0.0098) &   \\
    6 & DEBERTA & MDC & 0.5970 & 0.5988 & 0.5913 & 0.6031 & 0.5970 & 3 \\
       &         &     & (0.0041) & (0.0130) & (0.0092) & (0.0048) & (0.0041) &   \\
    7 & BERT & TIC & 0.6765 & 0.6688 & 0.6472 & 0.6664 & 0.6765 & 3 \\
       &      &     & (0.0067) & (0.0321) & (0.0260) & (0.0307) & (0.0067) &   \\
    8 & MPNET & TIC & 0.6825 & 0.6755 & 0.6559 & 0.6672 & 0.6825 & 3 \\
       &       &     & (0.0069) & (0.0218) & (0.0297) & (0.0324) & (0.0069) &   \\
    9 & DEBERTA & TIC & 0.6835 & 0.6810 & 0.6609 & 0.6656 & 0.6835 & 3 \\
       &         &     & (0.0024) & (0.0164) & (0.0239) & (0.0269) & (0.0024) &   \\
    \bottomrule
  \end{tabular}
  \caption{Recipient gender classification results for same-domain evaluation. The standard deviation for each metric, calculated across three different seeds, is shown below the corresponding value.
  }
  \label{tab:recipient-performance-detail}
\end{table}

\begin{figure}[H]
    \centering
    \includegraphics[scale=0.35]{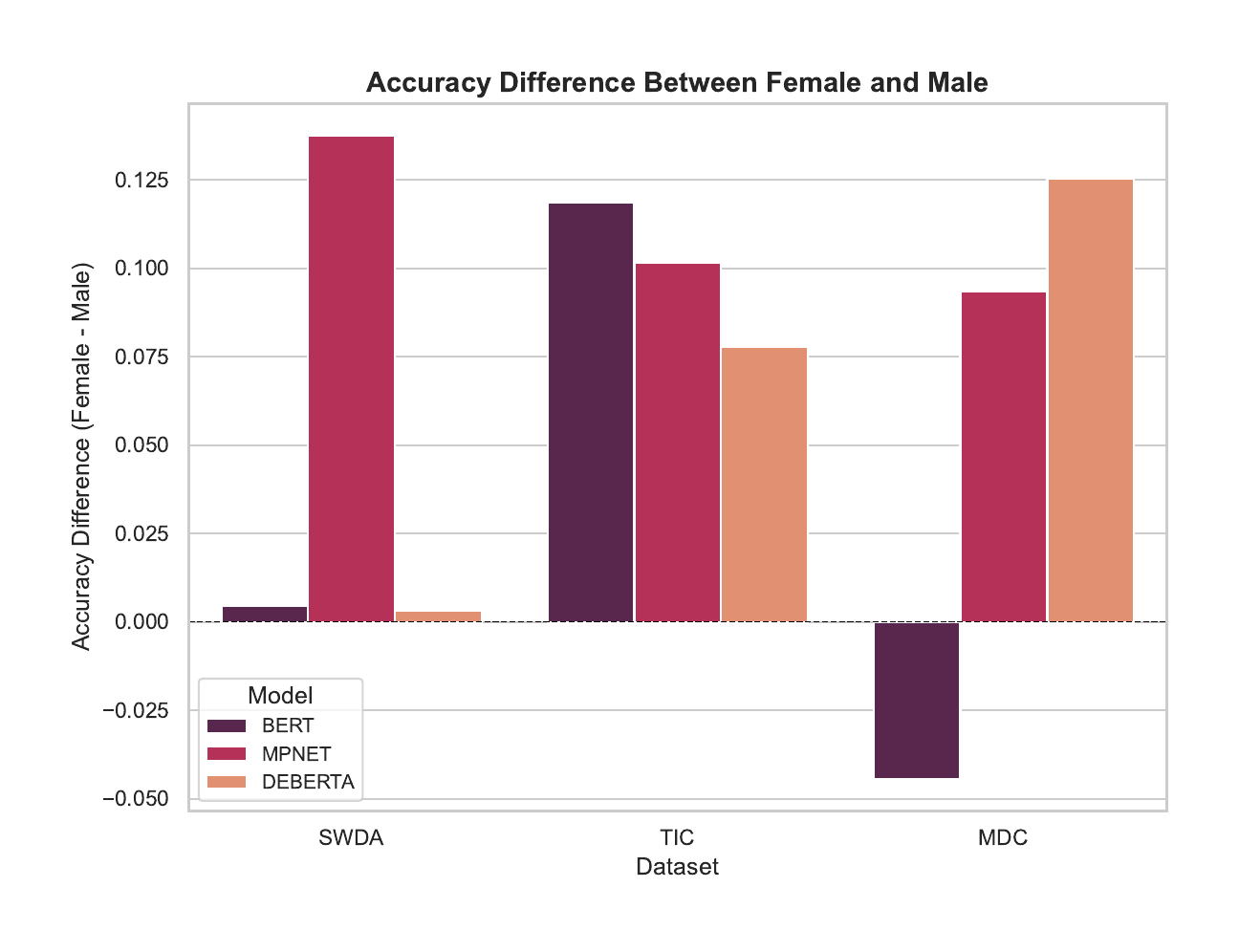}
    \caption{Difference in accuracy of fine-tuned models when predicting recipient gender. The values represent the average accuracy across three seeds. The models were train and test with in the same domain.}
    \label{fig:gender-diff}
\end{figure}

\section{Models Agreement Details} \label{sec:models-agreement}

In order to identify whether the models are capturing the same patterns, we compare their level of agreement on correct predictions. Let $P_{i,j}$ be the probability that models $i$ and $j$ agree, and $R_{i,j}$ be the probability of a random agreement between $i$ and $j$. The kappa coefficient between $i$ and $j$ can then be computed as:
\begin{equation}
\label{eq:kappa}
    \kappa_{i,j} = \frac{P_{i,j} - R_{i,j}}{1 - R_{i,j}}.
\end{equation}
Larger values of $\kappa_{i,j}$ indicates more agreement between the models $i$ and $j$ while negative values indicate disagreement.

\begin{figure}[H]
    \centering
    \includegraphics[width=1\linewidth]{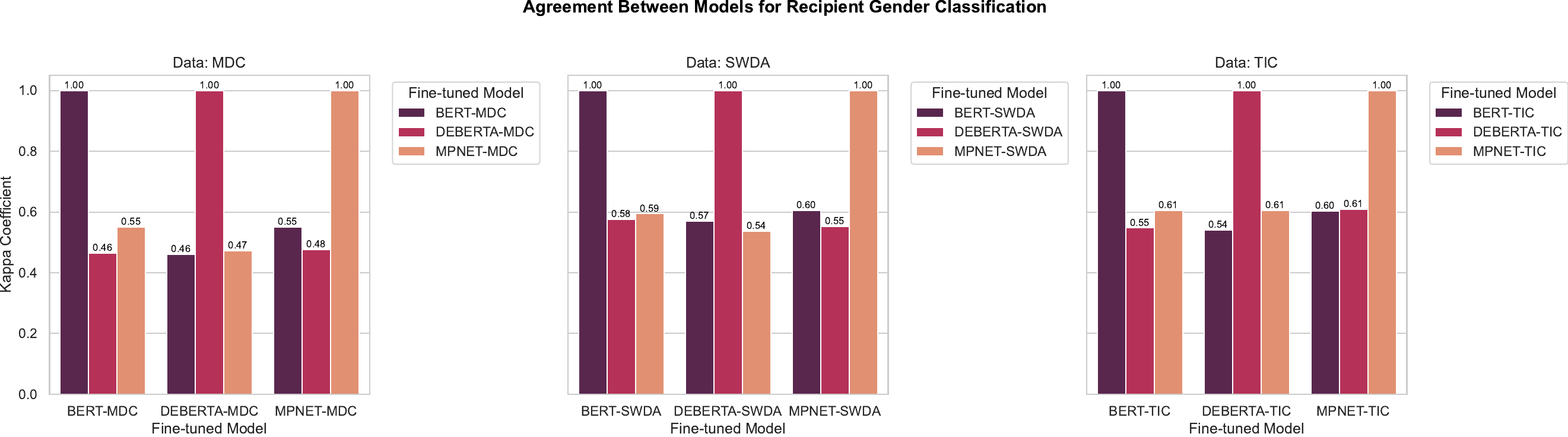}
    \caption{Agreement between fine-tuned models for recipient profiling, measured using the Kappa Coefficient to account for random agreement, as defined in Equation (\ref{eq:kappa}). The fine-tuned models were evaluated within the same domain. Each agreement value represents the average coefficient across three seeds used in the experiments. The agreement values for different models range from 0.46 to 0.61.}
    \label{fig:kappa-agreement}
\end{figure}
\end{document}